# Distance Metric-Based Learning with Interpolated Latent Features for Location Classification in Endoscopy Image and Video

Mohammad Reza Mohebbian*[1], Khan A. Wahid[1], Anh Dinh[1], and Paul Babyn[2]

[1]Department of Electrical and Computer Engineering, University of Saskatchewan, Saskatoon, Saskatchewan, S7N 5A9, Canada

[2]Saskatchewan Health Authority, Saskatoon, SK S7K 0M7, Canada.

## Abstract

Conventional Endoscopy (CE) and Wireless Capsule Endoscopy (WCE) are known tools for diagnosing gastrointestinal (GI) tract disorders. Detecting the anatomical location of GI tract can help clinicians to determine a more appropriate treatment plan, can reduce repetitive endoscopy and is important in drug-delivery. There are few research that address detecting anatomical location of WCE and CE images using classification, mainly because of difficulty in collecting data and anotating them. In this study, we present a few-shot learning method based on distance metric learning which combines transfer-learning and manifold mixup scheme for localizing endoscopy frames and can be trained on few samples. The manifold mixup process improves few-shot learning by increasing the number of training epochs while reducing overfitting, as well as providing more accurate decision boundaries. A dataset is collected from 10 different anatomical positions of human GI tract. Two models were trained using only 78 CE and 27 WCE annotated frames to predict the location of 25700 and 1825 video frames from CE and WCE, respectively. In addition, we performed subjective evaluation using nine gastroenterologists to show the necessaity of having an AI system for localization. Various ablation studies and interpretations are performed to show the importance of each step, such effect of transfer-learning approach, and impact of manifold mixup on performance. The proposed method is also compared with various methods trained on categorical cross-entropy loss and produced better results which show that proposed method has potential to be used for endoscopy image classification.

## 1. INTRODUCTION

Esophageal, stomach and colorectal tumors constitute about 2.8 million reported diagnoses and 1.8 million deaths annually around the world [1]. Endoscopy is considered the gold standard for gastrointestinal (GI) examination [2], and is key to early mucosal disease identification. All conventional endoscopy (CE) approaches, such as colonoscopy and gastroscopy, are invasive and may cause discomfort or patient harm [3]; however, they allow real-time video inspection and visualization of many gastrointestinal abnormalities, including esophagitis, polyposis syndromes, or ulcerative colitis [2]. On the other hand, the Wireless Capsule Endoscopy (WCE) offers a non-invasive means of GI inspection to scan areas that are inaccessible to conventional endoscopy such as the small bowel. A huge number of recorded frames need to be examined by an expert working at the workstation for diagnosis. However according to the literature [4], the diagnostic performance by visual inspection is low. For example, the diagnostic accuracy is about 69% for angioectasia, 46% for polyps, and 17% for bleeding lesions.

Accurately localizing the anatomic position of an abnormality within the GI tract is another challenge that remained unsolved [5–7]. There are various benefits in detecting location from endoscopy image. Accurate determination of the tip of the endoscope in the gastrointestinal tract, and hence the position of an abnormality, is important when further follow-up or surgery is needed [8], and helpful to reduce repetitive endoscopy attempts, to provide targeted drug delivery [9], and for automatic endoscopy navigation [10]. Additionally [11,12], some diseases characteristically happen at specific locations in the GI tract. For example, dangerous bleeding usually occurs in stomach, small bowel or duodenum [12]. Hence, providing location-based frame reviewing can reduce examination time and human error in high-risk regions.

However, endoscopy frame localization is challenging and may benefit from computer-aided systems. Figure 1 provides an illustration of this challenge showing two similar looking frames, but one from the early part of the stomach (cardia) and another from the end part (Pylorus). Differentiating these two frames by visual inspection may be difficult due to the high similarity between the two frames.

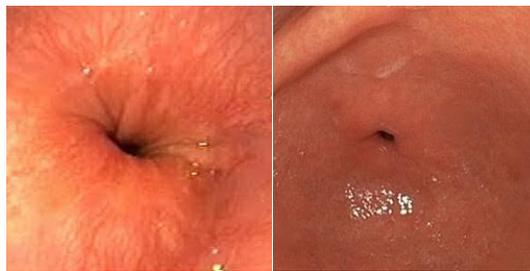

Figure. 1. Difficulty of detecting anatomical location form single image. The left image is for Cardia and the right image is for Pylorus

There are various methods used to localize endoscopy frames. Some methods perform localization using image processing techniques [13–25] while others use Radio Frequency (RF)-based approaches [26] or electromagnetic wave-based localization [27]. From methods that use image-processing, there are few studies that have used image classification [15,16,23–25]. Lee *et al.*[23] used the Hue-Saturation-Value (HSV) intensity variation in WCE video to recognize moving from one location to another location. They designed an event-based system to detect the esophagus, stomach, duodenal, ileum and colon (5 locations) and achieved 61% F1-score, however, they did not use any standard machine learning or deep learning approach as their method was based on heuristics. The combination of color features and support vector machine (SVM) is also performed by Marques *et al.* [15] for the stomach, small intestine, and large intestine (3 locations) classification on WCE frames. Their overall reported accuracy was 85.2% while the pylorus and ileocecal valve locations had the most error in the test set. Shen *et al.* [16] used the Scale Invariant Feature Transform for extracting local image features and the probabilistic latent semantic analysis model for unsupervised data clustering for localization of stomach, small intestine, and large intestine (3 locations) from WCE images. Esophagogastric junction, pylorus and ileocecal valve were distinguished as the most problematic parts and accuracy 99.9%, 98.3% and 94.7% reported for stomach, small intestine, and large intestine, respectively. Takiyama *et al*. were the first to use a convolutional neural network (CNN) for endoscopy location classification [25] using 27,335 standard endoscopy images training.. They classified larynx, esophagus, stomach (upper, medium, and lower part) and duodenum (6 locations) and could achieve 97% accuracy with AUC>99% on 13,048 images used in their test dataset. Finally, the Saito *et al.* [24] applied CNN on 4,100 standard colonoscopy images from the terminal ileum, cecum, ascending colon, transverse colon, descending colon, sigmoid colon, rectum, and anus (8 locations). They tested the model on 1,025 images and achieved 66% accuracy. All these methods were applied only on a limited number of locations and two studies were only applied on CE [24,27]. Additionally, the performance on lower GI locations still needs improvement.

For localization, other approaches used image processing but not image classification. Bao *et al.* [17] extracted color intensity, motion and texture features and used a kernel SVM for movement speed prediction and achieved 92.7% average accuracy for tracking WCE. Bao *et al.* [21] tried to localize the capsule with the aim of speed estimation in video frames. They analyzed consecutive frames to calculate the spatial displacement and achieved an average 93% accuracy for speed estimation and 2.49 cm error as localization error. Dimas *et al.* [20] proposed a novel visual odometry approach based on Multi-Layer Perceptron (MLP) which is applied on SIFT features. They estimated the WCE location according to some anatomic landmarks and reported an error of 2.70 ± 1.62 cm. As an illustration, the pylorus is considered as a starting point, and the displacement in the small intestine was calculated with respect to this point. Finally, in the context of RF localization, methods such as Received Signal Strength (RSS) and Time of Arrival (TOA) are used [26] and more recently, Shao *et al.* [27] introduced a passive magnetic localization method. Table 1 provides a brief background review of these endoscopic localization techniques.

TABLE I. A background of localization techniques that are used by others.

| REF | Method | Evaluation | Performance metrics and results |
|---|---|---|---|
| [23] | Variation in HSV intensity in subsequent frames using event correlation | Four anatomical locations classification: esophagus, stomach (entering stomach), small intestinal (entering duodenal and ileum), and colon | Recall: 76%; Precision: 51%; F1-score:61% |
| [25] | CNN | Six anatomical locations classification: larynx, esophagus, stomach (upper, medium, lower), duodenum | AUC: 100% for larynx and esophagus 99% for stomach and duodenum Accuracy: 97% |
| [24] | CNN | Seven anatomical locations classification: the terminal ileum, the cecum, ascending colon to transverse colon, descending colon to sigmoid colon, the rectum, the anus, and indistinguishable parts | AUC: 97% for the terminal ileum; 94% for the cecum; 87% for ascending colon to transverse colon; 84% for descending colon to sigmoid colon; 83% for the rectum; 99% for the anus. Accuracy: 66% |
| [13] | multivariate Gaussian classifiers with color, texture, motion features | Median error in frame number prediction for detecting esophagogastric junction; pylorus; ileocecal valve | Esophagogastric junction: 8 pylorus :91 ileocecal valve:285 (frames) |
| [14] | SVM with color and texture features | Median error in frame number prediction for detecting esophagogastric junction; pylorus; ileocecal valve | esophagogastric junction :2 pylorus: 287 ileocecal valve: 1057 (frames) |
| [22] | PCA and customized thresholding approach with color features | Median error in frame number prediction for detecting pylorus; ileocecal valve | Pylorus:105 ileocecal valve: 319 (frames) |
| [15] | SVM with color features | Three anatomical locations classification stomach, small intestine, and large intestine | 85.2 % (overall accuracy) |
| [16] | The probabilistic latent semantic analysis model for unsupervised data clustering with SIFT features | Three anatomical locations classification stomach, small intestine, and large intestine | stomach: 99.9% small intestine: 98.3% large intestine :94.7% Accuracy |
| [17] | kernel SVM with color intensity, motion, and texture features | Motion estimation is evaluated based on median error for detecting pylorus and ileocecal valve | 92.7% (average accuracy) |
| [21] | Feature Points Matching for capsule speed estimation | Speed estimation accuracy and location error | 93% accuracy for speed estimation and 2.49 cm for localization error |
| [20] | with SIFT features matched using random sample consensus and tracked using Kanade-Lucas-Tomasi tracker | robotic-assisted setup provided for evaluation | 2.70 ± 1.62 cm localization error |
| [26] | Using RSS, DoA or ToA | average RMSE for predicting capsule location | ≈100 mm RMSE with 10 sensors on body surface |
| [27] | Adding small magnet in capsule | Capsule inside a volume of 380 mm by 270 mm by 240 mm covered by 16 digital magnetic sensors | 10 mm RMSE error |

DoA: Directional of Arrival; ToA: Time of Arrival; PCA: principal component analysis; SIFT: Scale Invariant Feature Transform; SVM: support vector machine;

MLP: Multi-Layer Perceptron; CNN: Convolutional Neural Network; Hue-Saturation-Value: HSV; RMSE: root mean square error.

There is no available dataset that cover most anatomical landmarks for WCE and CE. Therefore, all previous works are applied for predicting limited number of locations. Moreover, all previous works are specialized for WCE or CE. The recent advances in AI can help to design models with least training samples to predict anatomical locations with high accuracy. Deep learning methods have yielded great results in image classification [28]. However, algorithm accuracy is highly dependent on training and typically requiring a large number of labelled datasets with a balanced number of samples per class. On the other hand, human visual systems can distinguish new classes with very few labelled instances [29]. The few shot learning (FSL) technique attempts to distinguish new visual categories from few labelled samples [30]. However, they suffer from overfitting issue because of low training samples. Introducing manifold mixup scheme could help models to have better decision boundaries between classes, while reducing overfitting possibility due to increasing number of training epochs [31].

In this paper, we designed a distance metric-based algorithm for extracting feature and localizing WCE and CE frame using few training samples for classifying 10 different anatomical locations. Since the number of training samples were few, the manifold mix-up scheme combined with few-shot learning model allowing us to increase the number of training epochs while decreasing the overfitting possibility. The manifold mixup also helped for making more precise decision boundaries. A subjective evaluation of anatomical location using images with nine gastroenterologists was initially conducted that shows that the performance of humans to identify GI location from images is poor. Therefore, an automated algorithm like the one proposed here is required to improve of the performance of GI diagnostic and frame localization.

## 2. Materials and methods

Two different models are created for CE and WCE frame localization based on Siamese Neural Network [32] (SNN) which is a type of few-shot learning. The model gets two images and calculates latent features for each image and compares these features using a distance loss. A manifold mix-up scheme is used to mix latent feature of images from support set to increase the number of training pairs and improve decision boundaries of the model. The final model is trained to predict distance of two input images. For using model in single frame localization, the input image is compared with all images from different locations and similarity to one group is determined using median of distances. Agreement of predictions among neighboring frames are used for localization of a frame sequence. More details are provided in the following sections.

### 2.1. Dataset collection

Two different datasets including images and videos have been used in this research. The image dataset consisted of both CE and WCE frames. It includes 78 CE and 27 WCE images from 10 different locations with at least 3 images in each class. The anatomical locations are depicted in Figure 2. Images were collected from the Gastrolab gallery [33] and a set of Pillcam images [34]. CE and WCE images were initially sized 256×256 and 512×512, pixels, respectively, which were resized to 256×256 pixels. Positions, including Esophagus, Cardia, Pylorus, Duodenum, Ileum, Jejunum and Colon (transverse, ascending, descending and sigmoid), had images for both CE and WCE. Only CE images were available for Rectum, Angularis, and Anus.

365 seconds of video captured by Pillcam and 1028 seconds of video captured by CE devices, were used to evaluate the performance. CE videos (25 frames per second) were taken from Gastrolab [33], and WCE videos (5 frames per sec) along with their annotations were taken from Faigel and Cave [34] book. Table II lists the datasets used in the work.

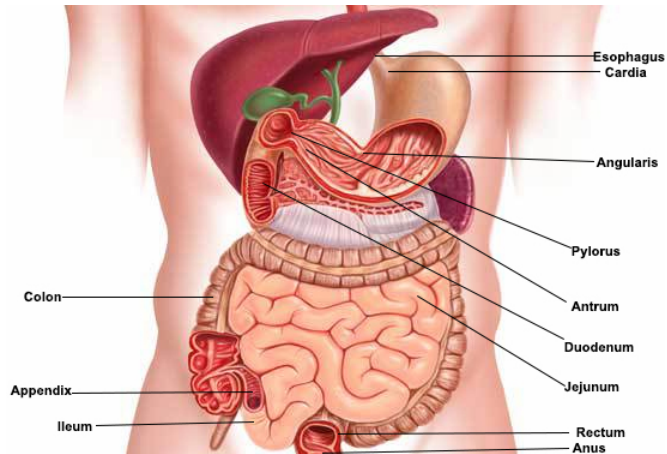

Figure. 2. The anatomical position of the images in dataset for human GI tract

TABLE II. Description of data set used for training and testing.

| Position | | Images (support set) | | Video frames | |
|---|---|---|---|---|---|
| Index | Name | CE | WCE | CE | WCE |
| 1 | Esophagus | 6 | 3 | 3075 | 260 |
| 2 | Cardia | 6 | 3 | 2450 | 20 |
| 3 | Angularis | 8 | 0 | 500 | 0 |
| 4 | Pylorus | 5 | 3 | 2500 | 280 |
| 5 | Duodenum | 16 | 5 | 2700 | 130 |
| 6 | Jejunum | 5 | 3 | 1500 | 380 |
| 7 | Ileum | 11 | 5 | 475 | 280 |
| 8 | Colon | 11 | 5 | 5400 | 475 |
| 9 | Rectum | 5 | 0 | 5100 | 0 |
| 10 | Anus | 5 | 0 | 2000 | 0 |
| Total (Frame) | | 78 | 27 | 25700 | 1825 |
| Total (Second) | | - | - | 1028 | 365 |

To determine the efficacy of the proposed method under real conditions, data are supplemented by numerous diseases. Half of these images in WCE and CE image-based dataset contain pathology, including polyps, vascular anomalies, cancer, and inflammation. On the other hand, the video-based dataset has about 6500 and 600 frames with abnormalities for CE and WCE, respectively.

## 2.2. SUBJECTIVE EVALUATION BY GASTROENTEROLOGISTS

We conducted a survey where nine gastroenterologists were asked to identify the anatomical location of 50 images from the image-based CE dataset. Figure 3 shows a screenshot of the questionnaire, which is also available on the website (https://human-endoscopy-localization.web.app).

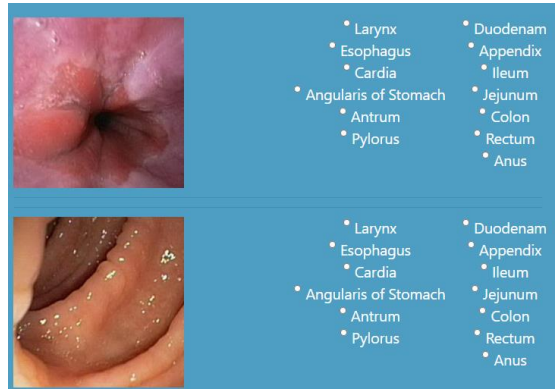

Figure 3. Two examples of survey questions used in the subjective evaluation. The questionnaire is available at *https://human-endoscopy-localization.web.app*

This CE dataset (Table II) contains frames from Esophagus, Cardia, Angularis of stomach, Pylorus, Duodenum, Ileum, Jejunum, Colon, Rectum and Anus. The responses of the gastroenterologists were later analyzed and F1-score, Accuracy, and area under the curve of ROC (AUC) are calculated. The objective was to evaluate the performance of the proposed AI-enabled system compared to diagnosis by visual inspection and show that an automated AI-based system can improve the diagnostic accuracy.

### 2.3. PROPOSED METHOD

#### 2.3.1. SIAMESE NEURAL NETWORK WITH MANIFOLD MIX-UP

FSL algorithms can be categorized into three major categories: initialization based, hallucination-based and distance metric learning based approaches. In initialization-based methods, the system focuses on learning to fine-tune or by learning an optimizer. The LSTM-based meta-learner to replace the stochastic gradient descent optimizer [29] is an example of this category. The hallucination-based approach tries to train a generator to augment data for a new class, and usually is used in combination with other FSL approaches such as distance-based method [35].

By learning to compare inputs, distance metric learning addresses the FSL problem. The hypothesis is that if a model can assess similarities between two images, it can identify an unknown input image. A distance-based classification model achieves competitive results with respect to other complex algorithms [28]. Siamese Neural Network (SNN) is an example of distance metric-based methods. SNN was first presented by Bromley *et al.* [32] in order to detect forged signatures. In that study, by comparing two signatures, the SNN was able to demonstrate whether two signatures were original or whether one was fake.

The FSL method proposed here, is a combination of a SNN using DenseNet121 with manifold mix-up scheme for having more training samples and better decision boundaries. The block diagram of the SNN is shown in Figure 4. The model is based on the extraction of two parallel latent features which have similar weights. Various deep learning approach can be used for feature extraction. The result of the network should be a feature vector (latent vector) for each image, which is usually a dense layer before last activation function. We tried different transfer learning approaches, which all were pre-trained on ImageNet [36], including DenseNet121, GoogleNet, AlexNet, Resnet50 and VGG16. DenseNet121 was selected for the baseline model since it showed the highest accuracy.

In the next step, the Euclidean distance between the two feature vectors is calculated after a linear transformation (Dense layer with size 64) and normalization. If both images are from the same class, the model learns to extract features that have less distance. On the other hand, if the two images come from separate groups, then the algorithm aims to obtain features that will make the distance greater. The sigmoid function is used to map the distance to range 0 to 1. This helps to have a better comparison of distances and helps manifold mix-up to have confined values [37].

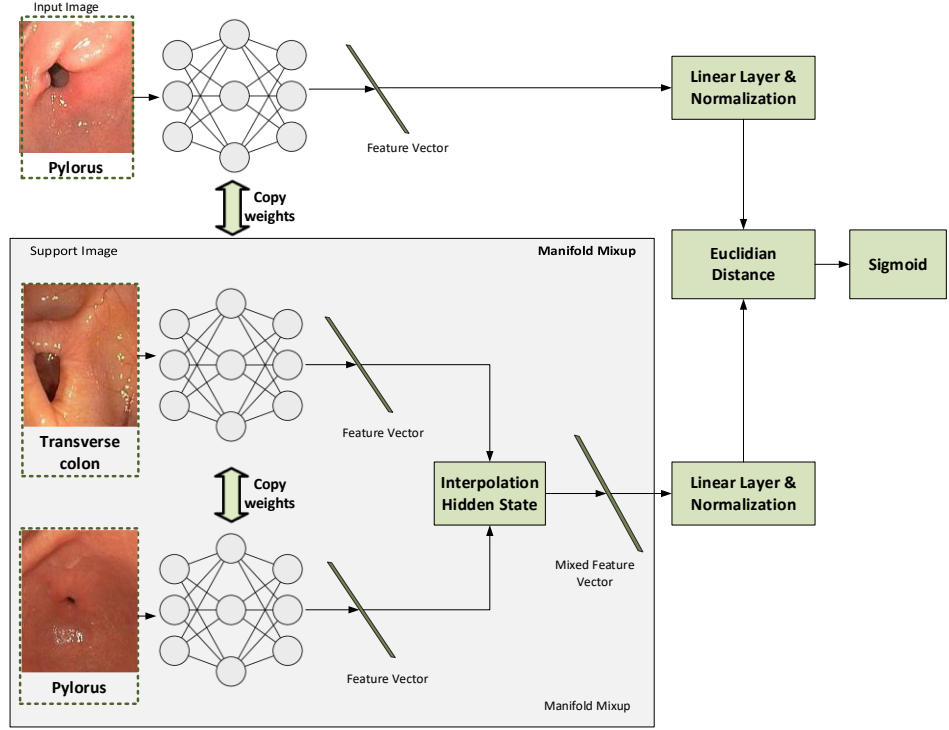

Figure. 4. The block diagram of the training SNN. Instead of using support set image directly, the mixing up of latent features is used for training.

The Contrastive loss is used to train the network [38]. The map for converting image to latent vector should preserve neighboring relationships and should be generalized on unseen data. The loss is defined as equation 1 below:

$$L(D_w, Y) = (1 - Y)D_w^2 + Y\{max(0, 1 - D_w)\}^2 \quad (1)$$

Where, $Y$ is 0 when $\overline{X_1}$ and $\overline{X_2}$ are similar and is 1 when they are different; $D_w$ is the Euclidean distance. The loss function is optimized using an RMSprop optimizer [39].

Deep learning networks usually perform appropriately on the data distribution they were trained on; however, they provide incorrect (and sometimes very confident) answers when evaluated on points from outside the training distribution; the adversarial examples are an example of this issue [40]. Manifold mix-up, introduced by Verma *et al.* [31], brings a regularization that solves this problem by training the classifier with interpolated latent features allowing it to be less confident at points outside of distribution. It enhances the latent representations and decision boundaries of neural networks. We suppose that extracted features from one location is unique to that location. As a result, combining latent features from two locations generates a new feature that is close to both locations, and the degree of resemblance is determined by the mixing weights.

Suppose $\check{x} = g(x)$ is the neural network function that maps one support image x to latent feature $\check{x}$. We assume two support images $x_1$ and $x_2$ and mix two latent features $\check{x}_1$ and $\check{x}_2$. The mixing function is defined by the following equation:

$$Mix_\lambda(\check{x}_1, \check{x}_2) = \lambda\check{x}_1 + (1 - \lambda)\check{x}_2 \quad (2)$$

Where, $\lambda$ is defined based on the $Betta(\alpha, \alpha)$ distribution [41] and $\alpha$ is set to 2 because the original paper achieved best result with this value. The bigger $\lambda$ means that the latent feature is more like $x_1$. Similarly, the labels of two support images $x_1$ and $x_2$, which are defined as $y_1$ and $y_2$, are mixed:

$$Mix_\lambda(\breve{y}_1, \breve{y}_2) = \lambda\breve{y}_1 + (1 - \lambda)\breve{y}_2 \qquad (3)$$

If two support images are in different locations than the input image of SNN network the output does not change. Therefore, one of the images should be from same location of input SNN. For each two pair, 50 different mixed latent features and labels are generated.

### 2.3.2. APPLYING MODEL TO A SINGLE FRAME AND A SEQUENCE OF FRAMES

Figure 5 shows the way for applying a single image to the trained model. When a new image is fed to the trained model, a feature vector is calculated. The Euclidean distance between obtained feature vectors and other classes are calculated; the minimum median distance from each group shows inclusion of the new image to a particular group. If the median distance from all group members is above the threshold of 0.5, a new category is generated for the image, and subsequently labeled as "Other". We used the median, instead of the average, which makes the algorithm more robust against noise [42].

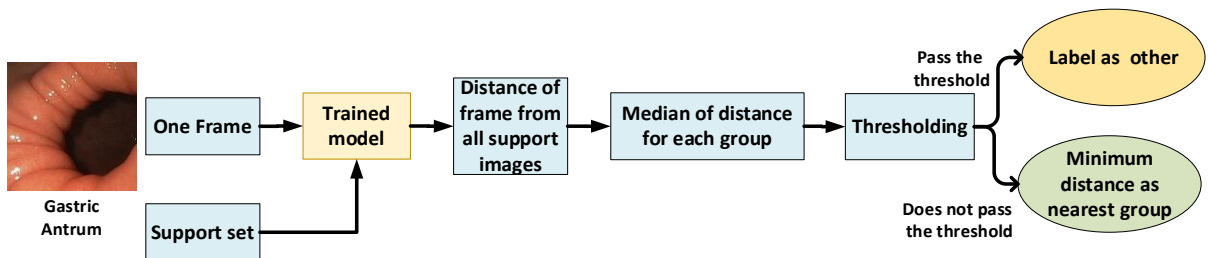

Figure. 5. The pipeline of applying a single frame to SNN for detecting location.

Figure 6 shows the block diagram of applying the model to a video sequence. Each video is segmented into 1 sec window with 0.5 sec overlap. Since the anatomic changes in video frames are not usually high, frames inside a window can be assigned to a location instead of assigning a location to each frame. Therefore, the error of applying model to a single frame, can be reduced by taking the advantage of temporal information. In this regard, each frame is applied to the single frame model. Then, the statistical mode of 1 second of frames location is used as the label of that second. It is worth noting that WCE and CE videos are in 5 and 25 frames per second.

Besides, it is assumed that the positions are in anatomical order, and the order should be preserved throughout the processing of a video sequence. For example, it is not possible for "Colon" to precede "Cardia". Hence, if the predicted label for sliding window was not ordered according to their anatomical positions, the label with higher average distance from its group is set to "Other".

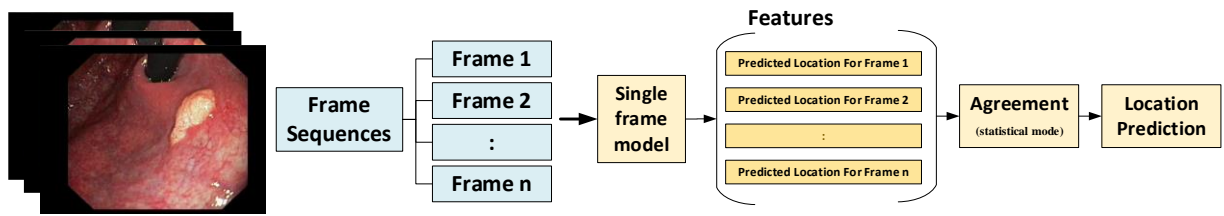

Figure. 6. The block diagram of applying trained model for predicting location of a frame sequence, which is applied on video-based dataset based on extracted features by SNN.

**2.4. DEFINING ABLATION STUDIES AND INTERPRETATIONS**

Deep learning ablation studies are based on the concept of ablation studies in neuroscience to explore the structure of information embodied by the network [43]. The idea is that certain parameters of a trained network contribute very little or none to the network's performance, making them insignificant and therefore able to be removed. We want to use this ablation approach not to improve the size and speed of a neural network but want to acquire insights into impact of each step on the performance, resulting in an interpretable model.

In the first ablation study, the effect of base model is evaluated. In this regard, the DenseNet121 is replaced with other transfer learning approach and the results are compared. In the next ablation study the advantage of distance metric base learning is investigated. Because of having a large number of video frames, one may wonder if traditional machine learning and deep learning techniques could be trained instead of the proposed approach. It is noteworthy to point out that the classification performance of the standard deep-learning models depends primarily on the sample size. Therefore, it is crucial to examine the amount of training data required to obtain a certain level of accuracy. In general, sample size should be at least relatively large compared to [44]; where is number of layers and is input dimension. Therefore, to have a three layers deep learning model with $256 \times 256$ pixels image size, 432 samples are required for each class. However, the temporal variations in video frames are not high enough to get this number of samples for each class, especially for WCE videos, wherein the capsule moves very slowly with low frame rate. Moreover, there are many frames in the videos that can be considered as outliers since no information can be extracted due to bubbles, instrument noise, blurring, contrast issues, color saturation, and other artifacts. Furthermore, some classes are absent in the video frames and therefore the test data is highly imbalanced.

Nevertheless, video-based dataset is divided to 50-50% as training and test set for training other methods. This means that 12,850 CE and 912 WCE images, are used for training. Different methods, including a CNN model, SVM with Scale Invariant Feature Transform (SIFT) features, SVM with color and texture features, transfer learning based on GoogleNet, AlexNet, Resnet50 and VGG16 which are pre-trained on ImageNet [36], and used for comparing with the proposed method. The second step of postprocessing which is a rule of preserving anatomical order is also applied after predicting a location of frame.

The color texture features are extracted using local binary pattern (LBP) approach [13] and radial basis function (RBF) with the help of Wu and Wang [45] method was used to set the soft-margin and RBF kernel parameters. The SIFT features are also extracted based on Dimas et al. [20] work. The proposed CNN model consists of two convolutional layers with 32 and 16 filters and $3 \times 3$ kernel size, and two dense layers with 32 and 13 units which are connected to a softmax layer for predicting the class number and is optimized using Nesterov Adam optimizer [46] on categorical cross entropy loss function. The reason that transfer learning is utilized is that these networks are pre-trained on a large dataset and having imbalance and low number of training sample may have less effect on transfer learning approach than other traditional technique [47].

Besides, the impact of manifold mixup scheme on performance is also investigated. For this purpose, the SNN without manifold mixup is trained and compared with the proposed method. The effectiveness of manifold mixup is evaluated based on feature vector visualization using model interpretation techniques.

Model interpretation refers to ways that humans can use to understand the behavior and expectations of a systems [48]. To understand what latent features the model is extracting from images, two different approaches are taken. First, the heatmap from the last layer of base model is calculated. Since DenseNet121 is used as base model, the last Batch Normalization layer, which has the shape of (16, 16, 1024) for single image in batch, is the last layer before latent feature vector. The latent feature vector is created based on GlobalMaxPooling layer. In other words, pixels that have maximum values are selected in feature vector. These locations show the most important regions of an image that the network used to calculate the feature vector.

t-Distributed Stochastic Neighbor Embedding (t-SNE) is a dimensionality reduction technique that is ideally suited for the visualization of high-dimensional data [49]. Besides of heatmap, the extracted latent feature from the model is also visualized using t-SNE for better interpretation of the trained model. All test samples are feed into the base model and the t-SNE of the latent features are calculated and depicted with and without manifold mixup.

### 2.5. PERFORMANCE EVALUATION

Two validations were applied on the proposed method. Firstly, we applied SNN for evaluating single frame model performance, which is tested on all frames from video-based dataset. For validating, the whole proposed system, including SNN and postprocessing, the test dataset was 50% of video-based dataset with was 13,762 endoscopy video frames.

From the standard summarization quality metrics, F1-score, Accuracy, and area under the curve of ROC (AUC) are used. F1-score, Accuracy and AUC for predicting location $i - th$ is described in equation 4, 5 and 6, respectively.

$$F1score_i = \frac{2 \times Recall_i \times Precision_i}{Recall_i + Precision_i} \qquad (4)$$

$$Accuracy_i = \frac{TP_i}{TP_i + TN_i + FN_i + FP_i} \qquad (5)$$

$$AUC_i = \frac{Recall_i + Specificity_i}{2} \qquad (6)$$

Wherein, precision, recall and specificity are defined as equations 7, 8, and 9:

$$Precision_i = \frac{TP_i}{TP_i + FP_i} \qquad (7)$$

$$True\ Positive\ Rate_i = Recall_i = \frac{TP_i}{TP_i + FN_i} \qquad (8)$$

$$1 - False\ Positive\ Rate_i = Specificity_i = \frac{TN_i}{TN_i + FP_i} \qquad (9)$$

The True Positive ($TP_i$) is the number of frames which belong to $i - th$ location and their locations are correctly identified. The False Positive ($FP_i$) is the number of frames does not belong to $i - th$ location but predicted as $i - th$ location; False Negative ($FN_i$) is the number of frames pertaining to $i - th$ location but missing in prediction and True Negative ($TN_i$) is the number of frames which belong to $i - th$ location but their locations are not correctly identified.

For multiclass problem, the macro-average of Accuracy, AUC and F1-score are reported. It is worth mentioning that the micro-average is not sensitive to individual group predictive results and can be misleading when data is imbalance [50]. The macro-average for F1-score and AUC is calculated like equation 4 and 6, with this change that the average of recall, precision and specificity for all classes are used. For multiclass problem, the overall accuracy is reported, which is the average of accuracy for all classes.

Specificity and recall are Type I and II errors, respectively, while F1-score and AUC are composite indices. Accuracy is skewed toward the majority class and is not a proper index when the data set is imbalanced (i.e., the prevalence rate is not about 0.5) [50]. When the prevalence is greater than 0.5, F1-score is also biased, and should be avoided. Therefore, AUC and ROC curve are used beside of F1-socre. All algorithms ran on a system with Core-i9, 16 GB of RAM, and 6 GB Graphic Cards NVIDIA GeForce GTX 1060 with Python 3.6 programming language.

## 3. RESULTS

### 3.1. SUBJECTIVE EVALUATION

The F1-score, Accuracy, AUC, and ROC curves of predicted locations by gastroenterologists is shown in Figure 7. It shows the macro-average F1-score, AUC and overall Accuracy to be 55%, 78% and 60%, respectively.

The numbers show that the task of localization of GI tract with only one frame is a difficult task. There are many similarities between different locations that may lead to more human error. The trade-off between recall and specificity is depicted by the ROC curve. Classifiers with curves that are closest to the top-left corner perform better. The ROC curves, Accuracy and F1-score show that the performance of expert is less in locations that are in the middle of GI tract. This is mainly because these locations are hard to be reached by conventional endoscopy and colonoscopy devices. Overall, the results suggest the need for an automated algorithm with higher accuracy as the one proposed in the work.

### 3.2. PROPOSED LOCALIZATION METHOD

The proposed SNN, which is trained on image-based dataset, is applied on video-based dataset for CE and WCE images without considering frame sequence and results are provided in Figure 8. Concisely, the proposed SNN method used DenseNet121 and trained on 78 CE images, could achieve macro-average F1-score and AUC and overall accuracy 78%, 90%, and 83%, respectively for CE. Similarly, the model trained on 27 WCE images and could achieve 78%, 90%, and 84% F1-score and AUC and overall Accuracy, respectively.

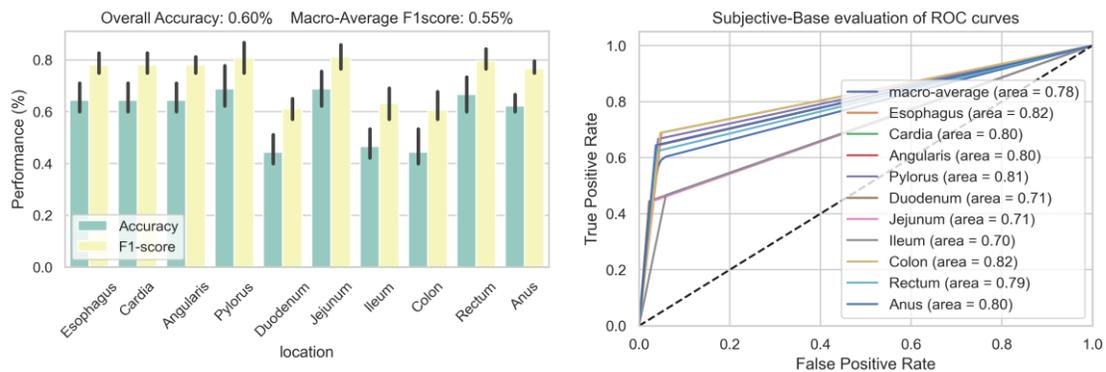

Figure 7. Performance of predicted location by nine gastroenterologists on CE dataset. The ROC curve for each location along with macro-average at the right and the F1-score and Accuracy are provided at the left.

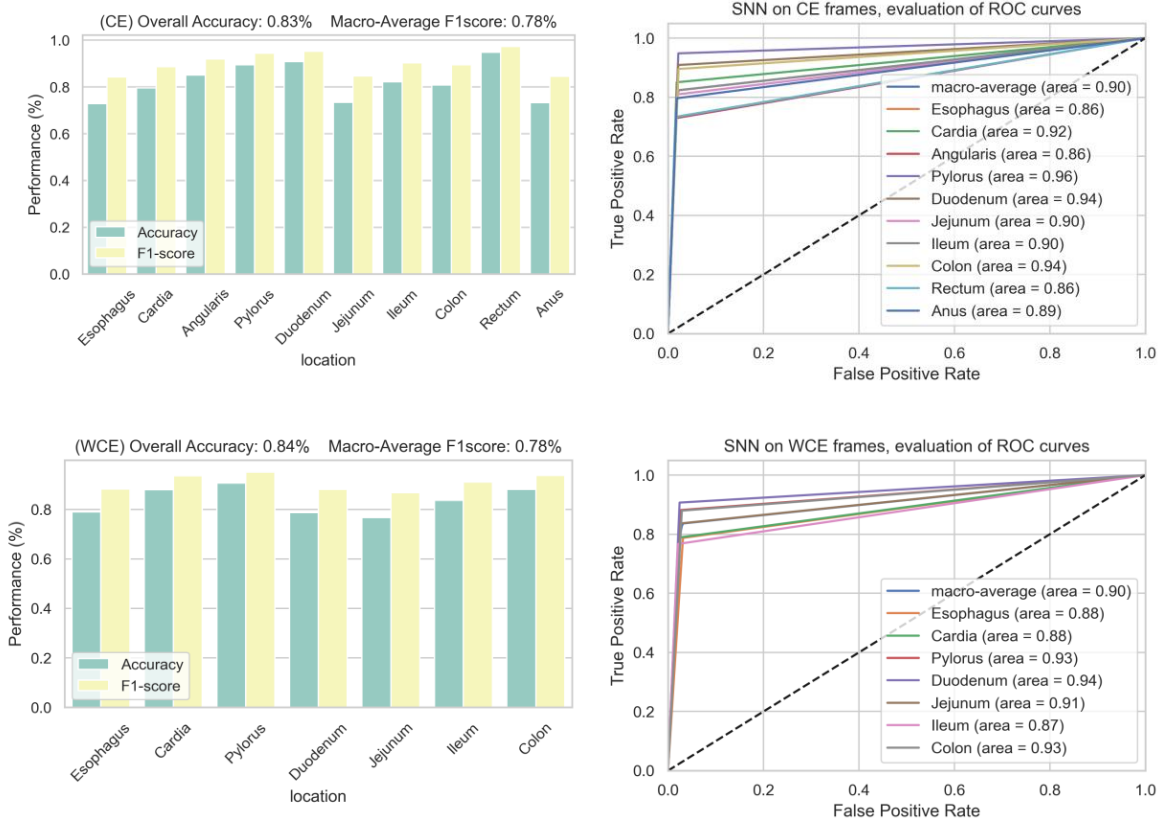

Figure 8. Results of SNN, trained on image-based dataset using manifold mixup, on single frames from CE (top) and WCE (bottom) video-based dataset.

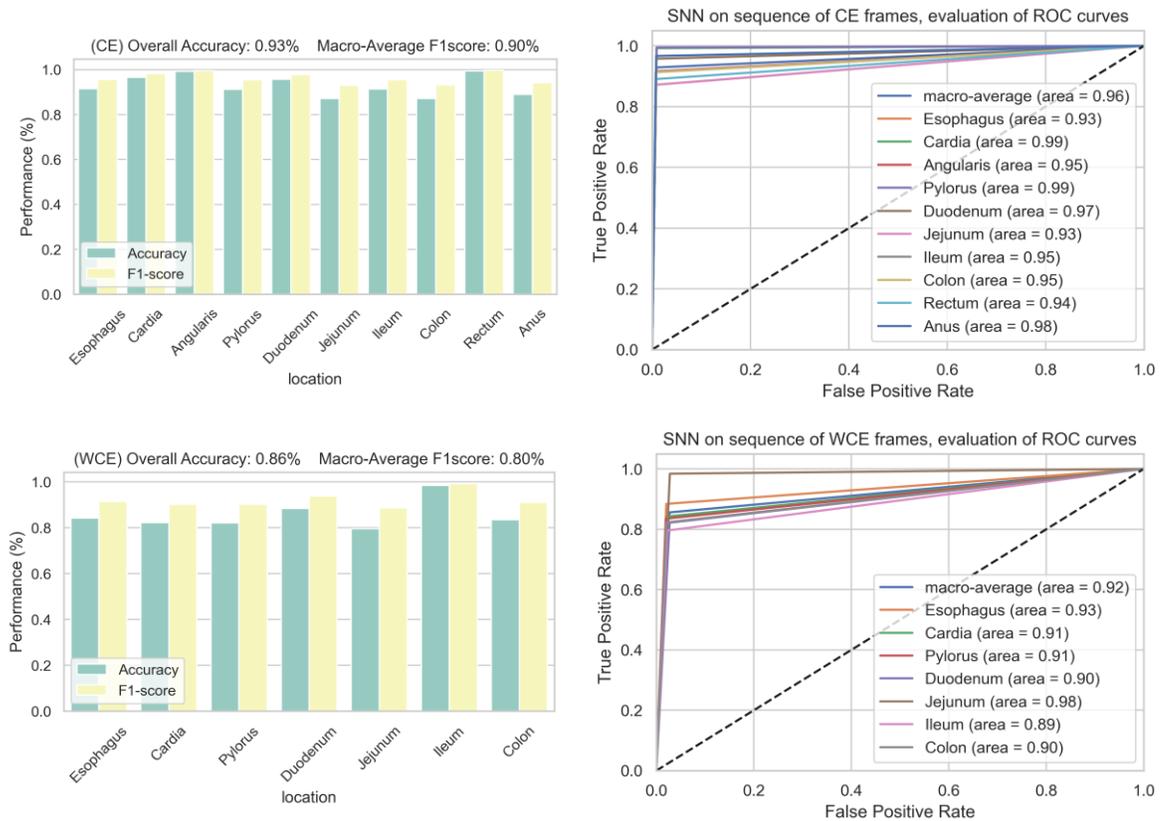

Figure 9. Results of SNN, trained on image-based dataset using manifold mixup, on sequence of frames from CE (top) and WCE (bottom) video-based dataset.

Figure 9 shows the effect of applying agreement (statistical mode) on sequence of frames. For using information from neighbor frames, the agreement of 25 and 5 frames were selected for CE and WCE location labels, respectively. The proposed method based on agreement of frame sequence predictions could achieve macro-average F1-score, AUC, and overall Accuracy 90%, 96%, and 93% for CE and 80%, 92%, and 86% for WCE, respectively.

An example of proposed method output is depicted on Figure 10 for processing a 34-second conventional endoscopy video. While the endoscope is in the Esophagus, there are times that proposed method (without agreement) cannot detect the correct location. Presenting different artifacts such as bubbles, instrument noise, blurring, contrast issues, color saturation, or simply that frame belongs to a location that was not in the train set such as antrum are examples of false predictions. The agreement of locations in a time frame can reduce error. As an instance, after detecting Esophagus position, the next positions, in this case Cardia, is expected to be predicted. Therefore, if irrelevant position is detected, the agreement process may fix the incorrect predicted frames.

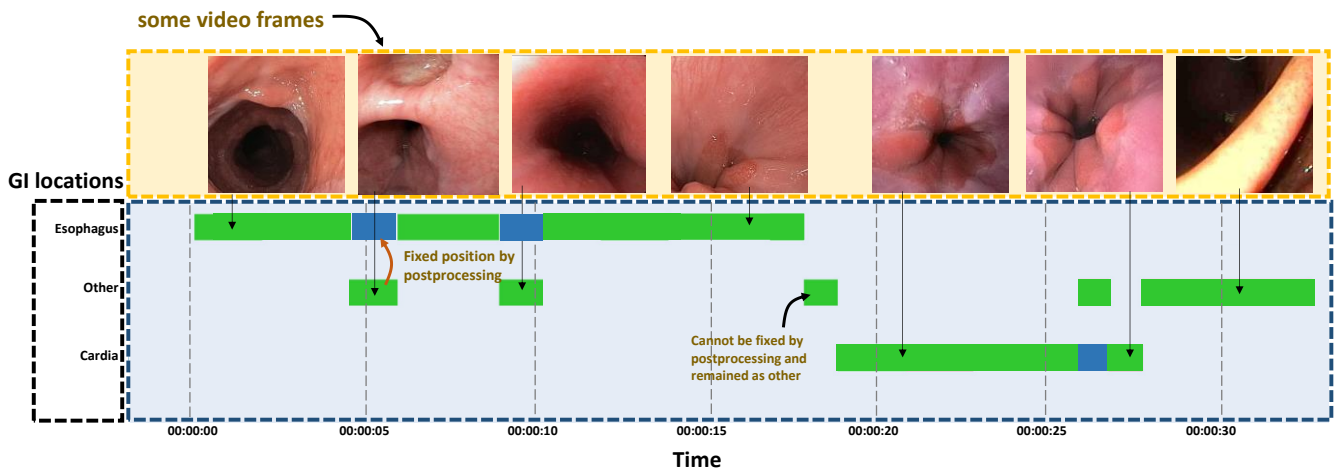

Figure 10. The overview the outputs of the system and error correction mechanism applied by our postprocessing step. "Other" label is mainly because of the inability of SNN to detect correct location because of artifact and noise, or it is a location that was not in the train set like Antrum. Blue boxes show erroneous predictions corrected using the proposed postprocessing step.

### 3.3. ABLATION STUDIES AND MODEL INTERPRETATION

#### 3.3.1. EFFECT OF CHANGING BASE MODEL

Instead of DenseNet121 that is used as the baseline model for getting feature vector, other transfer learning models, such as GoogleNet, AlexNet, Resnet50 and VGG16, which are pre-trained on ImageNet, are evaluated and DenseNet121 had a better performance. Figure 11 compares F1-score results of difference transfer learning approach. It shows that using other transfer learning approach for training models results in similar performance but the DenseNet121 is slightly better for current issue. As a result, other approaches may be used instead of DenseNet121 without a major performance difference.

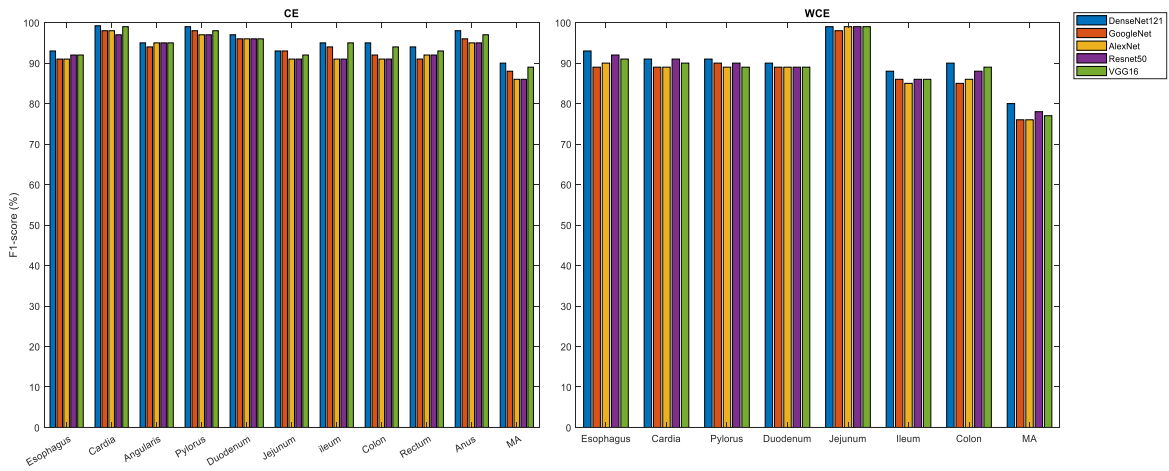

Figure 11. Comparing the F1-score of the proposed method with different transfer learning approach as base model. It shows that even though that DenseNet121 is selected for proposed method, other transfer learning approach can be used as based model without major performance difference.

### 3.3.2. EFFECT OF DISTANCE METRIC-BASED AND MANIFOLD MIXUP

Figure 12 shows the comparison results among SNN with manifold mix-up (proposed method), SNN without manifold mix-up, simple CNN, SVM with Scale Invariant Feature Transform (SIFT) features, SVM with color and texture features, GoogleNet, AlexNet, Resnet50 and VGG16. It is worth mentioning that proposed method with manifold mix-up is trained on limited data, while others (even SNN without manifold mix-up) are trained on 50% of frames from video-based dataset. The proposed method outperforms other models, although it is trained on only 78 CE and 27 WCE images and other models are trained on 12850 and 912 CE and WCE images. For CE, the VGG16 achieved the best score after the proposed method with macro-average F1-score 77.1%. On the other hand, Resnet50 gained the best score for WCE after the proposed method with macro average F1-score 73.7%, respectively. Additional information about VGG16 and Resnet50 is provided in the supplementary material.

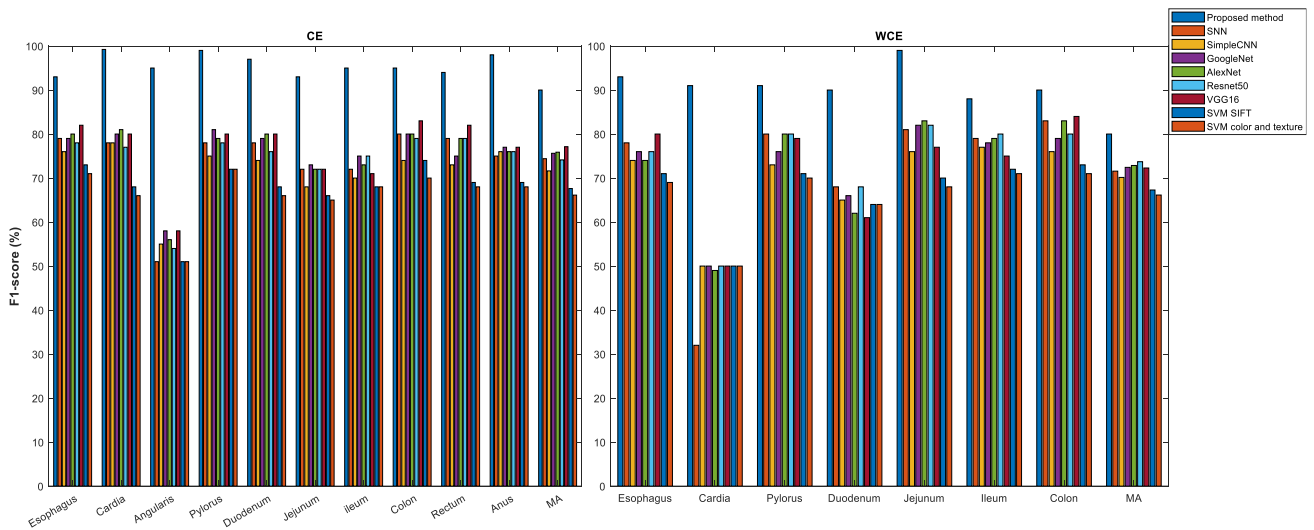

Figure 12. Comparing the F1-score of the proposed method using manifold mixup with SNN without manifold mixup and transfer learning classification and hand-crafted features with machine learning.

### 3.3.3. MODEL INTERPRETATION

For understanding the latent features extracted from images, the heatmap from the last layer of base model (DenseNet121) is provided in Figure 13. The heatmap should have places on (16, 16) matrix where maximum values appear (pink color). Besides, places where values change in different channels can also be informative about various features extracted by different filters (green color). The white color also shows the positions that have both maximum and standard deviation between channels. All the colored positions show the parts that model had attention toward it.

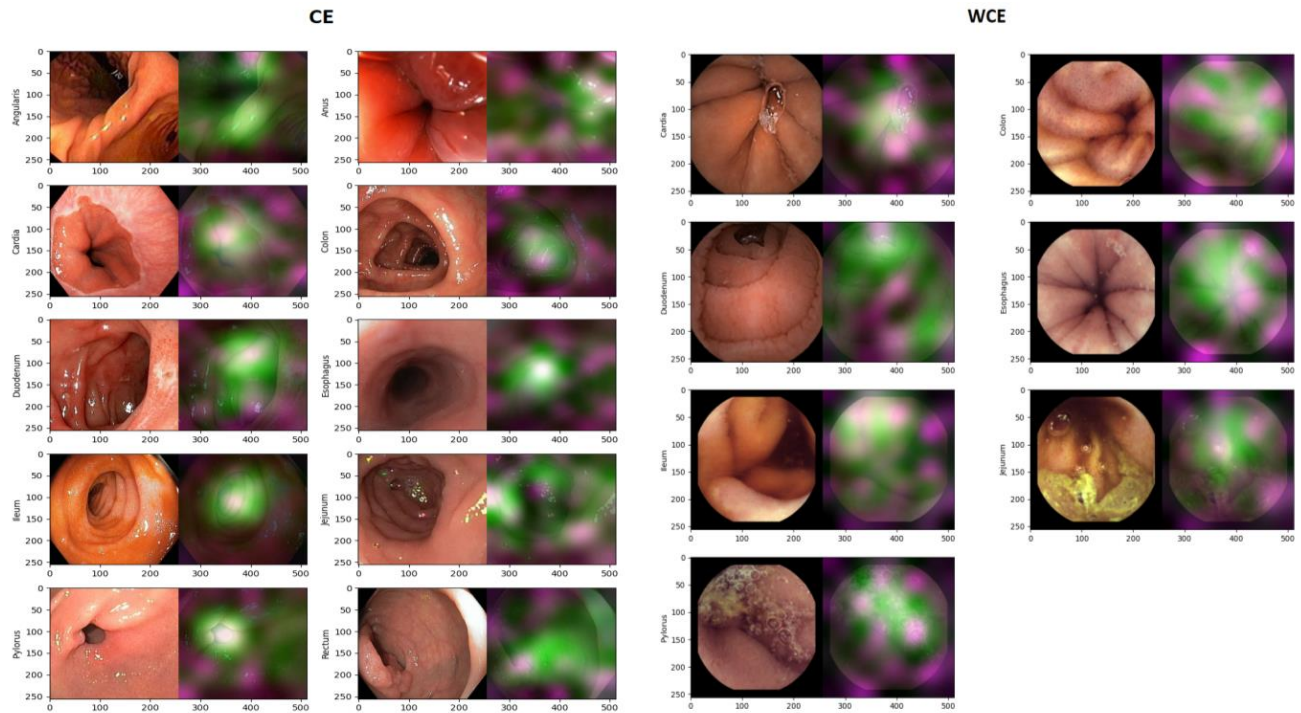

Figure 13. The heatmap from last layer of DenseNet121. The maximum values appear on filters of last layer are shown in pink color and high standard deviation pixels between all filters are shown in green. White regions are combination of pink and green parts.

It is challenging to explain how attention to these regions leads to distinguish between two different anatomical locations. Because the latent features acquired from these regions are passed through a linear transformation, then the distance is calculated. However, it is clear that the model gives emphasis regions, which could enable discriminant features to be extracted from those areas. For example, in a CE image of ileum, the model focuses on areas of the image that have more noticeable texture than others.

Figure 14 shows the latent vector visualization for CE and WCE images based on DenseNet121 on two dimensions using t-SNE. It is worth noting that since t-SNE holds probabilities rather than distances, calculating any error between Euclidean distances in high-D and low-D is pointless. Continuous lines in 2D plot also shows that there is a time series behavior in features, which is because of video frames. Moreover, the 2D plot shows that the complexity of manifold without Manifold mix-up scheme is higher (clusters are more correlated) and the manifold mix-up could help to find similarity better between frame sequences (more rigid lines).

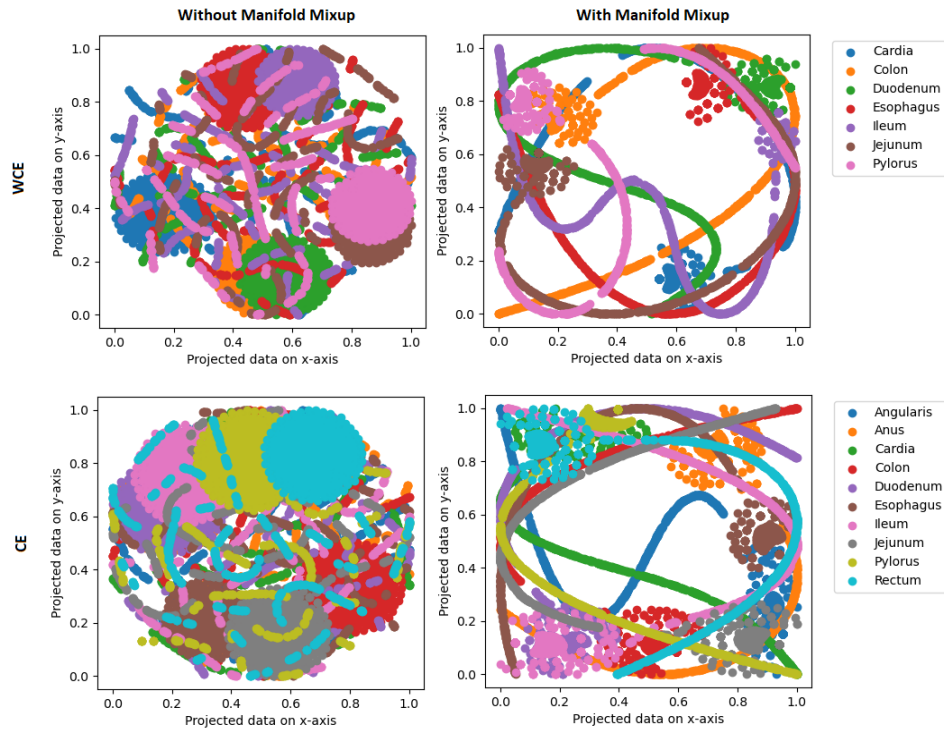

Figure 14. The visualization of latent feature extracted from CE and WCE video-based datasets using t-SNE with perplexity 50 based on proposed method with and without manifold mix-up. The latent features with manifold mixup have better discrimination; while the latent features extracted without manifold mixup have more overlaps.

## 4. DISCUSSION

In this paper, two SNN are trained using manifold mixup for localizing endoscopy on 78 CE and 27 WCE images. The trained systems are based on distance metric learning procedures, which can distinguish two images from different anatomical locations. Using frame sequence information, which is retrieved by agreement of predicted neighbor labels, the error rate is decreased.

### 4.1. PERFORMANCE AND PROPERTIES OF THE PROPOSED METHOD

As mentioned in the introduction, there are a limited number of studies that use image classification for endoscopy localization. Table 3 summarizes and compares these studies with our proposed method. Concisely, Lee *et al.*[23] designed a system to detect esophagus, stomach, duodenal, ileum and colon (5 locations) based on color change of videos and achieved 61% F1-score, however, they did not utilize any machine learning or deep learning approach. Marques *et al.* [15] used color features and SVM for the stomach, small intestine, and large intestine (3 locations) classification on WCE frames and achieved overall accuracy 85.2%. Shen *et al.* [16] used the SIFT local feature extraction on WCE images and unsupervised learning based on clustering for localization of stomach, small intestine, and large intestine (3 locations) and achieved overall accuracy of 97.6%. For the first time, Takiyama *et al* [25] used standard endoscopy images for training a CNN to classify input images into larynx, esophagus, stomach (upper, medium, and lower part) and duodenum (6 locations). They achieved 97% accuracy with AUC>99%. Next, the CNN is applied on standard colonoscopy images from terminal ileum, the cecum, the ascending colon, the transverse colon, the descending colon, the sigmoid colon, the rectum, and the anus (8 locations) by Saito *et al.* [24]. They achieved 66% overall accuracy.

All methods are applied on limited number of locations. However, in this research both WCE and CE localization are investigated with wide range of location from Esophagus to the Anus. Having more classes makes the problem more complicated. Increasing number of classes is also investigated in other fields such as anomaly detection. For instance, Mohammed *et al.* [51] showed that increasing number of classes makes the problem more complicated and causes drop in performance. On other hand, having more locations for prediction, makes the localization more precise. Furthermore, number of images that we used for training is significantly lower than other methods.

TABLE III. Comparing the performance and properties of the proposed method with other methods that used image classification as localization.

| STUDY | METHOD | ENDOSC OPY IMAGES | NUMBER OF LOCATIO NS | LOCATIONS | TRAUINING SAMPLE SIZE | VALIDATION STRATEGY (TEST SIZE) | BEST RESULT |
|---|---|---|---|---|---|---|---|
| [23] | Variation in HSV intensity in subsequent frames using event correlation | WCE | 5 | esophagus, stomach (entering stomach), small intestinal (entering duodenal and ileum), and colon | NA | External Dataset (10 videos, number of frames is NA) | Recall: 76%; Precision: 51%; 61% F1-score |
| [25] | Convolutional Neural Network | CE | 6 | Larynx, Esophagus, Stomach (Upper, Medium, Lower), Duodenum | 27335 | 13048 | 97% Accuracy |
| [24] | Convolutional Neural Network | CE | 6 | the terminal ileum, the cecum, ascending colon to transverse colon, descending colon to sigmoid colon, the rectum, the anus | 4100 | 1025 | 66% Accuracy |
| [15] | SVM with color features | WCE | 3 | stomach, small intestine, and large intestine | 26469 | 10-fold cross validation | 85 % Accuracy |
| [16] | The probabilistic latent semantic analysis model for unsupervised data clustering with Scale Invariant Feature Transform (SIFT) features | WCE | 3 | stomach, small intestine, and large intestine | 50000 | 10-fold cross validation | 97.6% Accuracy |
| PROPOSED METHOD | Attention-based SNN with Manifold mixup | WCE and CE | 10 for CE 7 for WCE | Esophagus, Cardia, Angularis, Pylorus, Duodenum, Jejunum, Ileum, Colon, Rectum, Anus | 78 CE 27 WCE | External Dataset (2570 CE, 1825 WCE) | CE: 93% Accuracy WCE: 86% F1-score |

NA: Not Available.

## 4.2. FUTURE WORKS AND LIMITATIONS

Having many labeled data including positions and abnormalities can help to design methods to diagnose the abnormalities along with localization; because some abnormalities may occur more on specific sites and this fact may help to improve the localization and anomaly detection.

Visualized t-SNE of features showed that there is a time series between latent feature of a video. This characteristic may help to design time series models based on recurrent neural network to process a video for localization. Furthermore, since attention and transformer are pioneers in autoregressive models, those type of architectures can be also used for processing a sequence of frames.

Although the performance of gastroenterologists on localization of single conventional endoscopy frame is assessed, there are more opportunity to expert performance on WCE frames and sequence of CE and WCE frames. Using frame sequence information can help the gastroenterologists to have better recognition about the location.

## 5. CONCLUSION

In this paper, a few-shot learning approach based on Siamese Neural Network and Manifold Mix-up is utilized to classify WCE and CE images based on their anatomical locations. The proposed method is only trained on 78 and 27 CE and WCE images, respectively. However, using the distance metric-based approach and manifold mix-up the number of training pairs are increased substantially which decreased the overfitting possibility. Moreover, the manifold mix-up scheme helped to have better decision boundaries and distance estimation. The proposed method is tested on external dataset, including 25,700 CE and 1825 WCE video frames, and achieves macro-average F1-score, AUC, and overall Accuracy of 90%, 96%, and 93% for CE and 80%, 92%, and 86% for WCE, respectively. Various ablation studies are carried out to demonstrate the significance of each part of the proposed method. The results of ablation studies showed that in the proposed method, other transfer learning models can also be used instead of DenseNet121 without major changes in performance. Moreover, it showed that the distance metric approach with manifold mixup, which are trained on few samples, have potential to outperform models which are trained using categorical cross-entropy loss on poorly sampled data. As instances, the proposed method outperformed other techniques, including a support vector machine with hand-crafted features, a convolutional neural network, and transfer learning-based methods which are trained on categorical cross-entropy loss. The visual inspection performed by nine experts on images also showed that an AI system can outperform visual inspections and it can help to improve diagnosis performance.

## ACKNOWLEDGMENT

We thank all physicians who completed the survey. Also, we would like to thank Natural Sciences and Engineering Research Council of Canada (NSERC) for supporting this work.